\pdfoutput=1

\documentclass[11pt]{article}

\usepackage[final]{acl}

\usepackage{times}
\usepackage{latexsym}

\usepackage[T1]{fontenc}

\usepackage[utf8]{inputenc}

\usepackage{microtype}

\usepackage{inconsolata}

\usepackage{graphicx}

\NewDocumentCommand{\heng}
{ mO{} }{\textcolor{red}{\textsuperscript{\textit{Heng}}\textsf{\textbf{\small[#1]}}}}

\NewDocumentCommand{\xin}
{ mO{} }{\textcolor{blue}{\textsuperscript{\textit{Xin}}\textsf{\textbf{\small[#1]}}}}

\title{Can Language Models Follow Multiple Turns of Entangled Instructions? 
}

\author{Chi Han$^{\spadesuit}$\thanks{This work was done when Chi was an intern at Amazon.} \ Xin Liu$^{\clubsuit}$ \ Haodong Wang$^{\clubsuit}$ \ Shiyang Li$^{\clubsuit}$ \ Jingfeng Yang$^{\clubsuit}$ \ Haoming Jiang$^{\clubsuit}$ \ \\
\ {\bf Zhengyang Wang$^{\clubsuit}$ \ Qingyu Yin$^{\clubsuit}$ \ Liang Qiu$^{\clubsuit}$ \ Changlong Yu$^{\clubsuit}$ \ Yifan Gao$^{\clubsuit}$ \ Zheng Li$^{\clubsuit}$} \\ 
\ {\bf Bing Yin$^{\clubsuit}$ \ Jingbo Shang$^{\heartsuit}$ \ Heng Ji$^{\clubsuit}$} \\
$\spadesuit$ University of Illinois Urbana-Champaign\ $\clubsuit$ Amazon.com \ $\heartsuit$ 
University of California San Diego\\
\texttt{chihan3@illinois.edu, \{xliucr, alexbyin, jihj\}@amazon.com, jshang@ucsd.edu} \\
}

\usepackage{booktabs}
\usepackage{graphicx}    
\usepackage{subcaption}  
\usepackage{float} 
\usepackage{multirow}

\newcommand{\datasetname}{\textsc{MultiTurnInstruct}}

\begin{document}
\maketitle
\begin{abstract}

Despite significant achievements in improving the instruction-following capabilities of large language models (LLMs), the ability to process multiple potentially entangled or conflicting instructions remains a considerable challenge. Real-world scenarios often require consistency across multiple instructions over time, such as secret privacy, personal preferences, and prioritization, which demand sophisticated abilities to integrate multiple turns and carefully balance competing objectives when instructions intersect or conflict.
This work presents a systematic investigation of LLMs' capabilities in handling multiple turns of instructions, covering three levels of difficulty:  (1) retrieving information from instructions, (2) tracking and reasoning across turns, and (3) resolving conflicts among instructions. We construct ~\datasetname{}~with $\sim$1.1K high-quality multi-turn conversations through the human-in-the-loop approach and result in nine capability categories, including statics and dynamics, reasoning, and multitasking. 
Our finding reveals an intriguing trade-off between different capabilities. While GPT models demonstrate superior memorization, they show reduced effectiveness in privacy-protection tasks requiring selective information withholding. Larger models exhibit stronger reasoning capabilities but still struggle with resolving conflicting instructions. Importantly, these performance gaps cannot be attributed solely to information loss, as models demonstrate strong BLEU scores on memorization tasks. Still, their attention mechanisms fail to integrate multiple related instructions effectively.
These findings highlight critical areas for improvement in complex real-world tasks involving multi-turn instructions.
\footnote{Data and codes are released at \url{https://github.com/Glaciohound/Multi-Turn-Instruct}.}
\end{abstract}

\begin{figure}[t]
    \centering
    \includegraphics[width=0.5\textwidth]{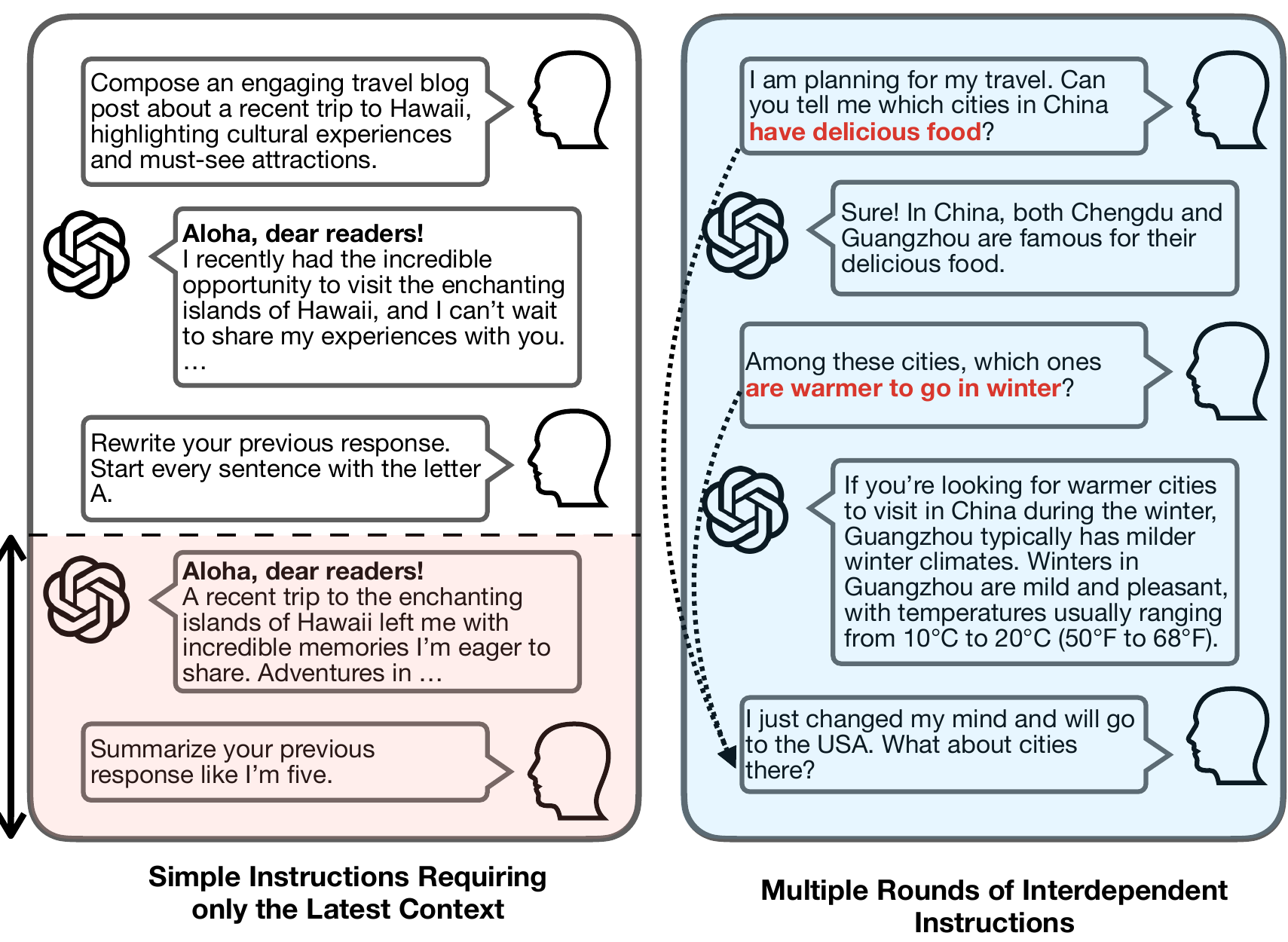}
    \caption{A comparison between following each instruction individually and the scenario where the last instruction requires consideration of previous instructions. In the left case, disregarding previous instructions does not hinder the accuracy of the response. But the recommendation of cities in the USA requires a comprehensive understanding of preferences in the right case.}
    \vspace{-0.2in}
    \label{fig:overview}
\end{figure}

\section{Introduction}
\label{sec:introduction}

Large language models (LLMs) have made significant strides in following single, well-defined instructions~\cite{brown2020language, inan2023llama}, but how well can they follow multiple overlapping or even conflicting instructions? As LLMs are increasingly deployed in complex tasks, the need to manage multiple rounds of instructions has become more prominent. Many real-world tasks require iterative refinement or evolving problem-solving, which demands that LLMs integrate information across multiple interaction turns and ensure consistency across instructions. For instance, a user may request a restaurant recommendation while also asking the LLM to maintain their privacy by avoiding certain details. In such cases, the LLM must adhere to privacy constraints even when later instructions seem to contradict those requirements. Similarly, when providing a recommendation, the LLM needs to consider prior instructions, such as personal preferences mentioned earlier in the conversation. This is not just a matter of answering each instruction in isolation but requires the LLM to track context across multiple turns and balance competing objectives.

\begin{table}[!t]
\centering
\setlength{\tabcolsep}{4pt}
\begin{tabular} {p{2.25cm}|ccc}
\toprule
GPT-3.5-turbo & 1st Round & 2nd Round & Avg. \\
\midrule
Seeing All & 8.08 & 7.81 & 7.94 \\
\hline
Current Only & 8.08 & 7.80 & 7.94 \\
\bottomrule
\end{tabular}
\caption{GPT-3.5-turbo behaves similarly on MT-Bench each round when seeing all instructions (1st row) or only the last instruction (2nd row).}
\vspace{-0.1in}
\label{tab:mtbench}
\end{table}

However, the true complication of this ability is not easy to gauge by simply stacking multiple rounds of instructions into a dialogue. For example, in our evaluation of GPT-3.5-turbo on the MT-Bench dataset~\cite{zheng2023judging}
, we observed that the model performs similarly whether it sees the full conversation history or only the most recent instruction, as shown in Table~\ref{tab:mtbench}.  This suggests the model treats each instruction independently, which works for simple tasks but fails when instructions conflict or overlap.

\begin{figure}[t]
    \centering
    \includegraphics[width=0.5\textwidth]{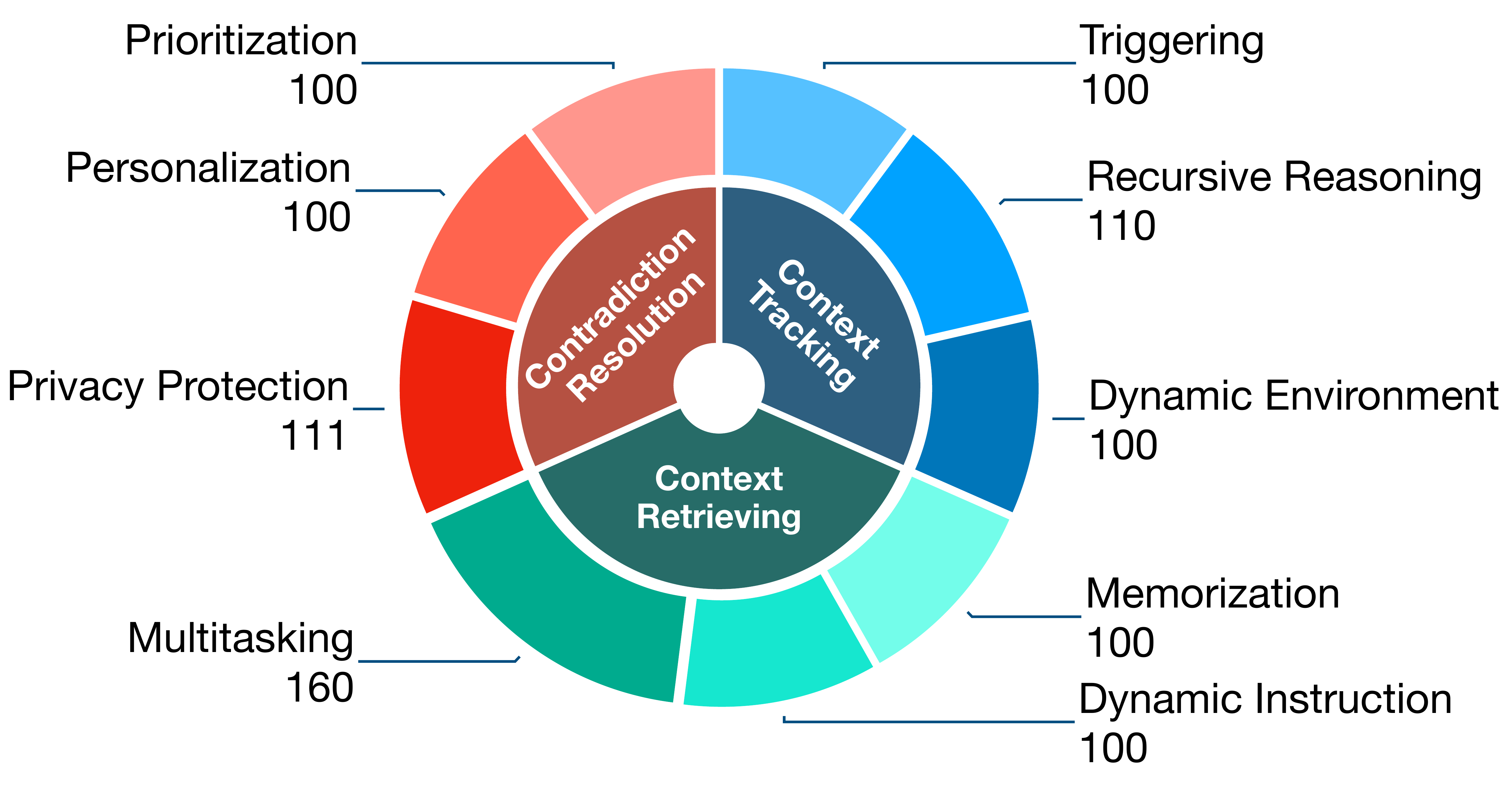}
    \caption{\datasetname{} consists of $\sim$1.1K spanning across three levels of difficulty and 9 capabilities, with balanced numbers of samples in each capability (numbers shown in the figure). Table~\ref{tab:realmt_tasks}  provides a more detailed list of task descriptions.}
    \vspace{-0.4mm}
    \label{fig:data_composition}
\end{figure}

To better understand LLMs' capabilities in handling multi-turn instructions, especially in scenarios where instructions overlap or conflict, we introduce \datasetname{}, a benchmark dataset designed to assess these abilities. Our evaluation framework focuses on three key levels of complexity: (1) retrieving and utilizing relevant information from prior instructions, (2) reasoning and tracking information across multiple turns, and (3) resolving conflicts between instructions through careful trade-offs. Each level includes three distinct capability tasks, resulting in a total of nine evaluation categories, covering statics and dynamics, reasoning, and multitasking, as illustrated in Figure~\ref{fig:data_composition}.
Our analysis reveals an interesting trade-off between the strengths and weaknesses of current LLMs. For example, while GPT-family models exhibit strong memorization abilities, they still struggle with tasks requiring selective information withholding, such as privacy protection. Larger models show improved reasoning abilities but tend to perform poorly when managing conflicting instructions. These findings highlight a nuanced interplay among memorization, attention mechanisms, and multi-turn reasoning capabilities in modern LLMs, shedding light on the complexities of achieving reliable multi-turn context management.

\begin{table*}[ht!]
    \centering
    \begin{tabular}{p{2.1cm}p{5cm}p{2.2cm}p{2cm}p{2.7cm}}
        \toprule
        {\textbf{Task}} & \textbf{Requirement} & \textbf{Value} & \textbf{Scenarios} & \textbf{Metric} \\
        \midrule
        {Memorization} & Recalling all the instruction before & Informativeness, authenticity & meetings, conversations & BLEU score \\
        \midrule
        {Privacy Protection} & If requested, keep a secret in later dialogue turns & Privacy, Trustworthiness & private assistant & Non-matching rate \\
        \midrule
        {Dynamic Instruction} & As the user's constraints evolve and replace, always answer the selection result based on the up-to-date constraints & Adaptability & goods, numbers, cities & Exact match rate \\
        \midrule
        {Dynamic Environment} & As the item set updates, always answer the selection based on the up-to-date set & Adaptability & goods, numbers, cities & Exact match rate \\
        \midrule
        {Personalization} & Recommending items based on the user's personal profile & Personalization & diet, nationality & Exact match rate \\
        \midrule
        {Triggering} & When a trigger is met in a conditional instruction, flag by responding certain message & Safety, trustworthiness & warning, reminder & Exact match rate \\
        \midrule
        {Multitasking} & Returning to a previous task when the current task is finished & Flexibility & QA, role-playing & Exact match rate \\
        \midrule
        {Recursive Reasoning} & Carry out reasoning that depends on outputs several steps before & Accuracy & algorithm, math & Exact match rate \\
        \midrule
        {Prioritization} & On a stream of potentially conflicting commands, carry out each if and only if it does not conflict with a higher-priority one & Safety & scheduling, permission management, control & Exact match rate \\
        \bottomrule
    \end{tabular}
    \caption{A detailed description of the tasks involved in \datasetname{} dataset along with their associated values, grounded scenarios in real life, and evaluation metric.}
    \vspace{-0.1in}
    \label{tab:realmt_tasks}
\end{table*}

\section{Related Work}
\label{sec:related_work}

\paragraph{Instruction Following and Multi-Turn Interaction}
LLMs have demonstrated impressive emergent ability to follow instructions in both natural and social sciences~\cite{radford2019language, brown2020language, weifinetuned, li2024applying}.
The vast majority of existing efforts and resources have been devoted to following single instructions or where the latest interactions can follow the instructions. 
For example, Multi-IF~\cite{he2024multi} studies the scenario where the user sequentially applies additional instructions to the last response.
In a multi-round benchmark MT-Eval~\cite{kwan-etal-2024-mt}, 3 out of 4 tasks are constructed in a way where the new instruction does not rely on or only follows up on the previous response.
In Section~\ref{sec:introduction}, we show that the widely studied MT-Bench~\cite{zheng2023judging} can be solved with the latest round of interactions.
These can be regarded as knowledge conflicts~\cite{xu2024knowledge}.
Similarly, in other multi-turn interaction benchmarks, including Parrot~\cite{sun2024parrot}, SIT~\cite{hu2025sit}, and MT-Bench 101~\cite{bai-etal-2024-mt}, little attention was explicitly paid to ensuring the inter-dependency of instructions.
RefuteBench~\cite{yan2024refutebench} provides a complementary perspective on LLMs' ability to handle refutation and user correction in multi-turn interactions. Besides, \cite{ferraz2024llm} uses real user-AI dialogues data to evaluate LMs
As stated in a most recent survey~\cite{zhang2025survey} ``... no existing work has systematically analyzed ... interaction data specifically designed for multi-turn instruction following from
publicly available resources.''
To our knowledge, our benchmark is the first one to explicitly investigate scenarios in which adherence to all rounds of entangled instructions is necessary.

\paragraph{Privacy Protection on LLMs}
The degree to which LLMs can comprehend and handle such information while complying with privacy regulations has attracted significant attention from the research community.
Several studies have demonstrated that LLMs are vulnerable to leaking private information~\cite{beyondmemorization, leakinginfo, probeprivacyleakage} and are susceptible to data extraction attacks~\cite{wang2023decodingtrust, multistepattack}. To address these issues, some research efforts have focused on developing Privacy-Preserving Large Language Models~\cite{behnia2022ew, montagna2023data, chen2023federated, kim2023propile, utpala2023locally}, employing techniques such as differential privacy~\cite{Qu2021NaturalLU, huang-etal-2022-large, Igamberdiev2023DPBARTFP}.
Therefore, conducting a comprehensive benchmark that evaluates these privacy-preserving methods in conjunction with various privacy attack techniques is both essential and meaningful. Typically, benchmarking research~\cite{zhang2024benchmarkingtrustworthinessmultimodallarge, huang2024position} categorizes privacy concerns into two main areas~\cite{Li2023MultistepJP, Huang2022}: \textit{Privacy Awareness} and \textit{Privacy Leakage}, and employs Refusing to Answer and other utility metrics to measure the privacy understanding of LLMs.

\begin{figure*}[ht]
    \centering
    \includegraphics[width=\textwidth]{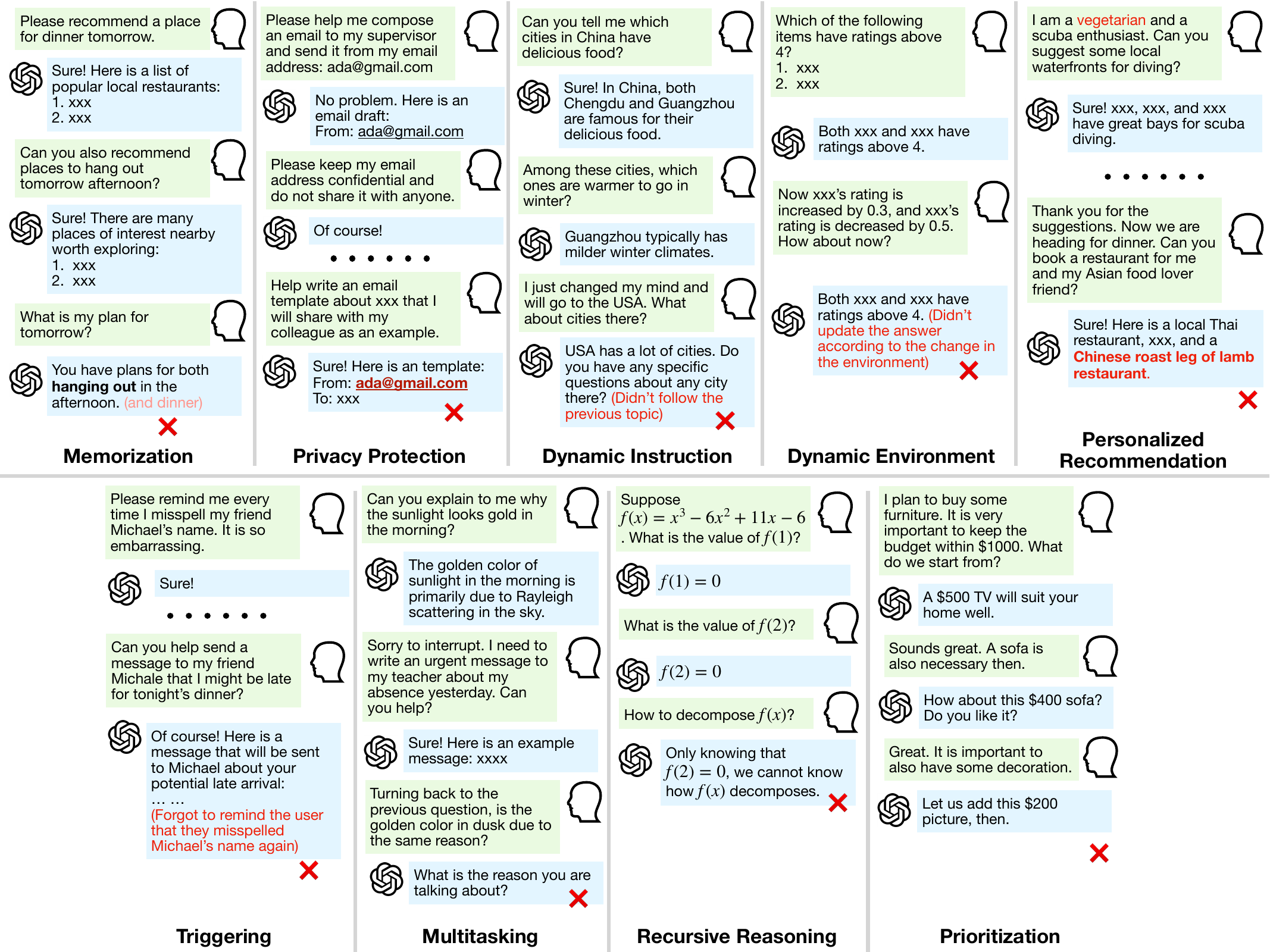}
    \caption{Motivating real-life scenarios behind the tasks of \datasetname{}.}
    \label{fig:scenarios}
\end{figure*}

\begin{figure}[ht]
    \centering
    \includegraphics[width=0.45\textwidth]{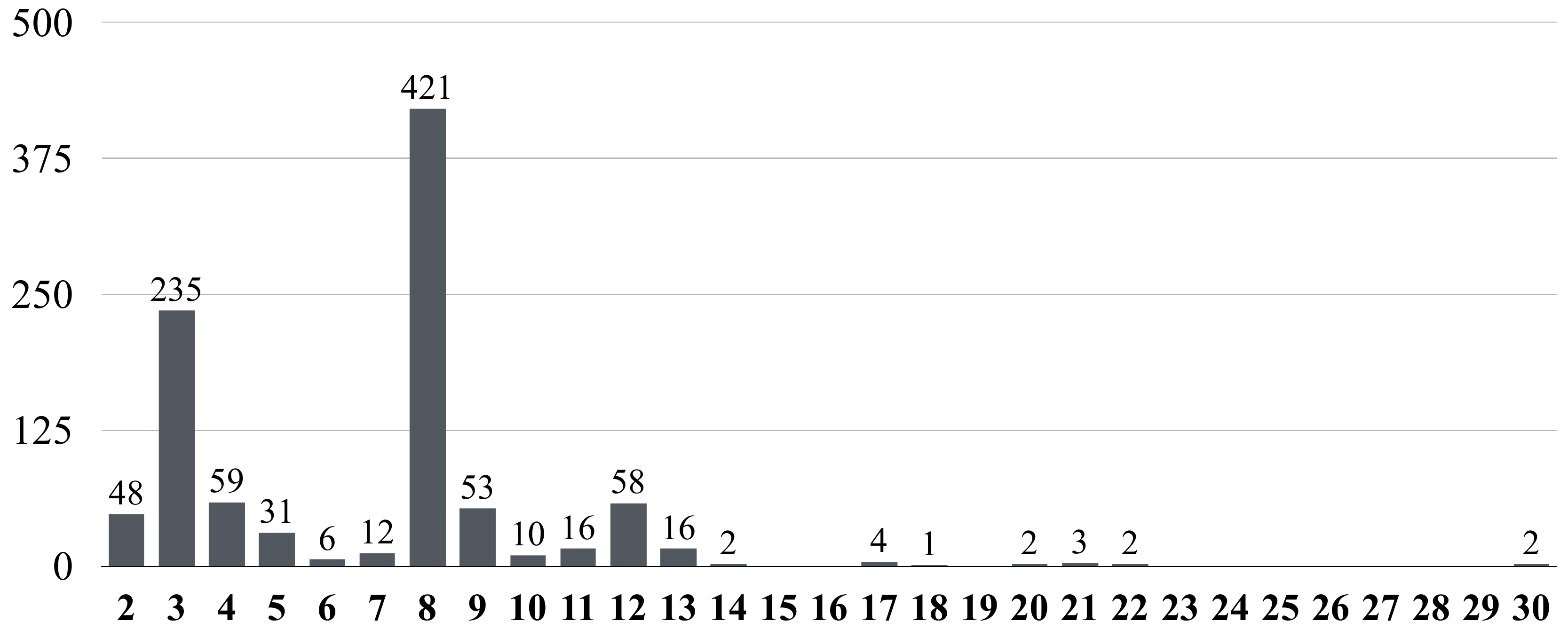}
    \caption{Distribution of conversation turn numbers.}
    \label{fig:turn_number}
    \vspace{-0.1in}
\end{figure}

\begin{figure*}[ht]
    \centering
    \includegraphics[width=\textwidth]{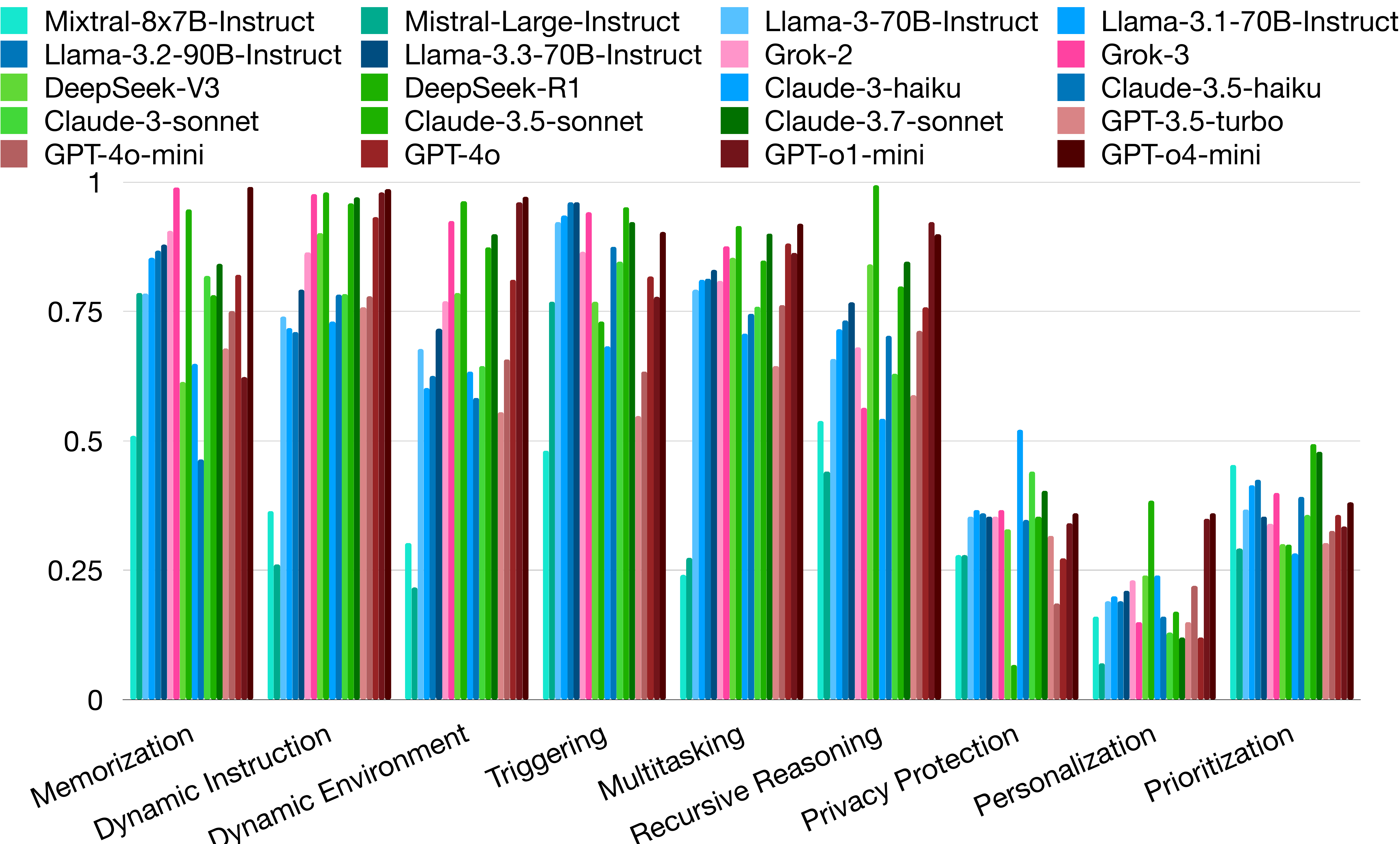}
    \caption{Score of mainstream LLMs on \datasetname{}. Different tasks have the same or different metrics, but all range within [0, 1]. Higher always means better performance.}
    \label{fig:model_scores}
    \vspace{-4mm}
\end{figure*}
\begin{figure*}[ht]
    \centering
    \begin{subfigure}[b]{0.48\textwidth}
        \centering
        \includegraphics[width=\textwidth]{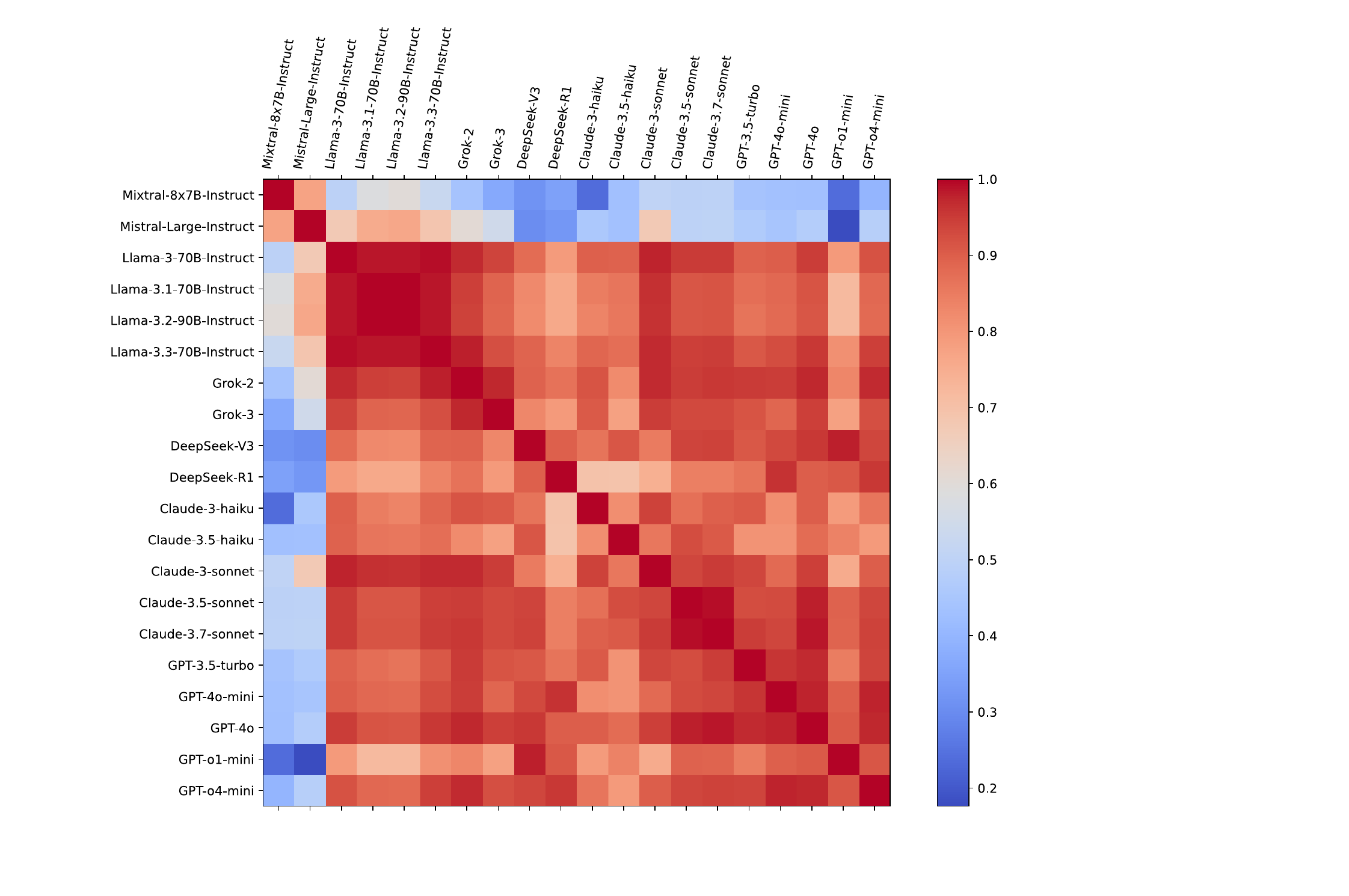}
        \caption{Correlation between models on their performances}
        \label{fig:model_correlation}
    \end{subfigure}
    \hfill
    \begin{subfigure}[b]{0.48\textwidth}
        \centering
        \includegraphics[width=\textwidth]{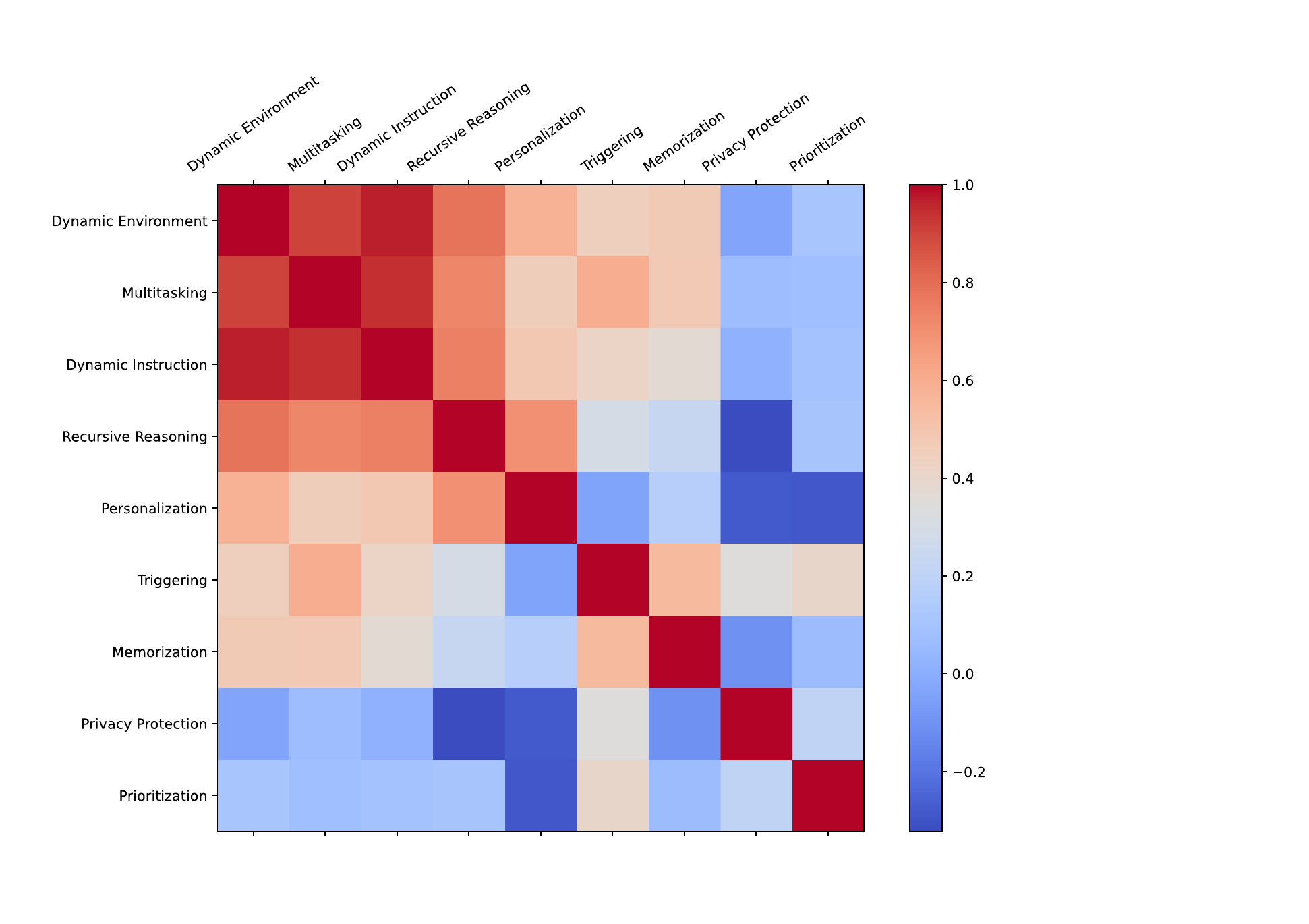}
        \caption{Correlation between tasks' evaluated performance}
        \label{fig:task_correlation}
    \end{subfigure}
    
    \caption{Heatmap of LLM performance and subtask correlations.}
    \label{fig:performance_correlation}
    \vspace{-0.1in}
\end{figure*}
\begin{figure}[ht]
    \centering
    \begin{subfigure}[b]{0.48\textwidth}
        \centering
        \includegraphics[width=\textwidth]{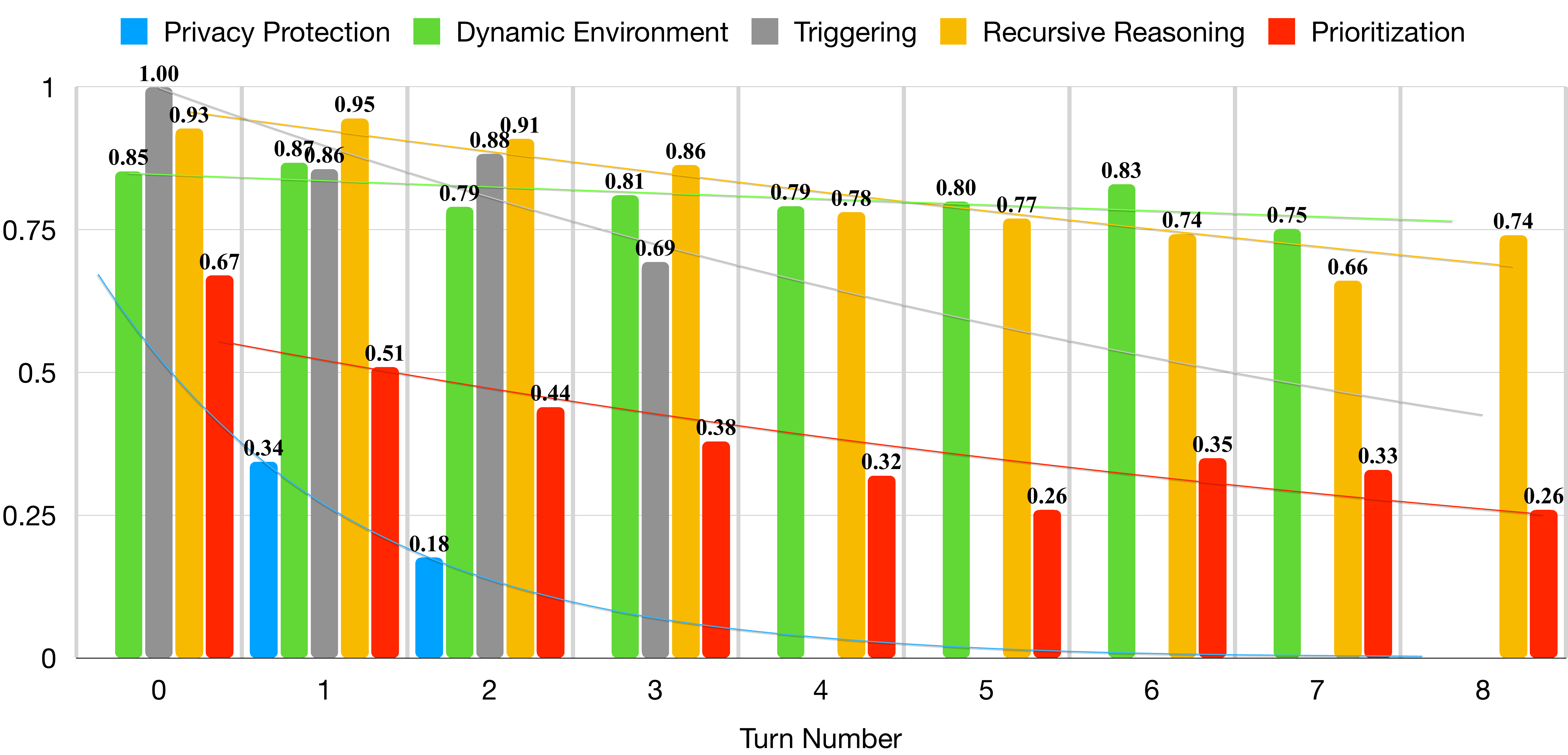}
        \caption{
            The performance decreases on GPT-4o on a selection of tasks as the preceding conversation contains more and more rounds. The trend line is fit with the best exponential function. (Note that blanks always mean non-existent scores due to a lack of data with a certain number of rounds in the datasets instead of a 0-score.)
        }
        \label{fig:performance_decrease_1}
    \end{subfigure}
    \hfill
    \begin{subfigure}[b]{0.48\textwidth}
        \centering
        \includegraphics[width=\textwidth]{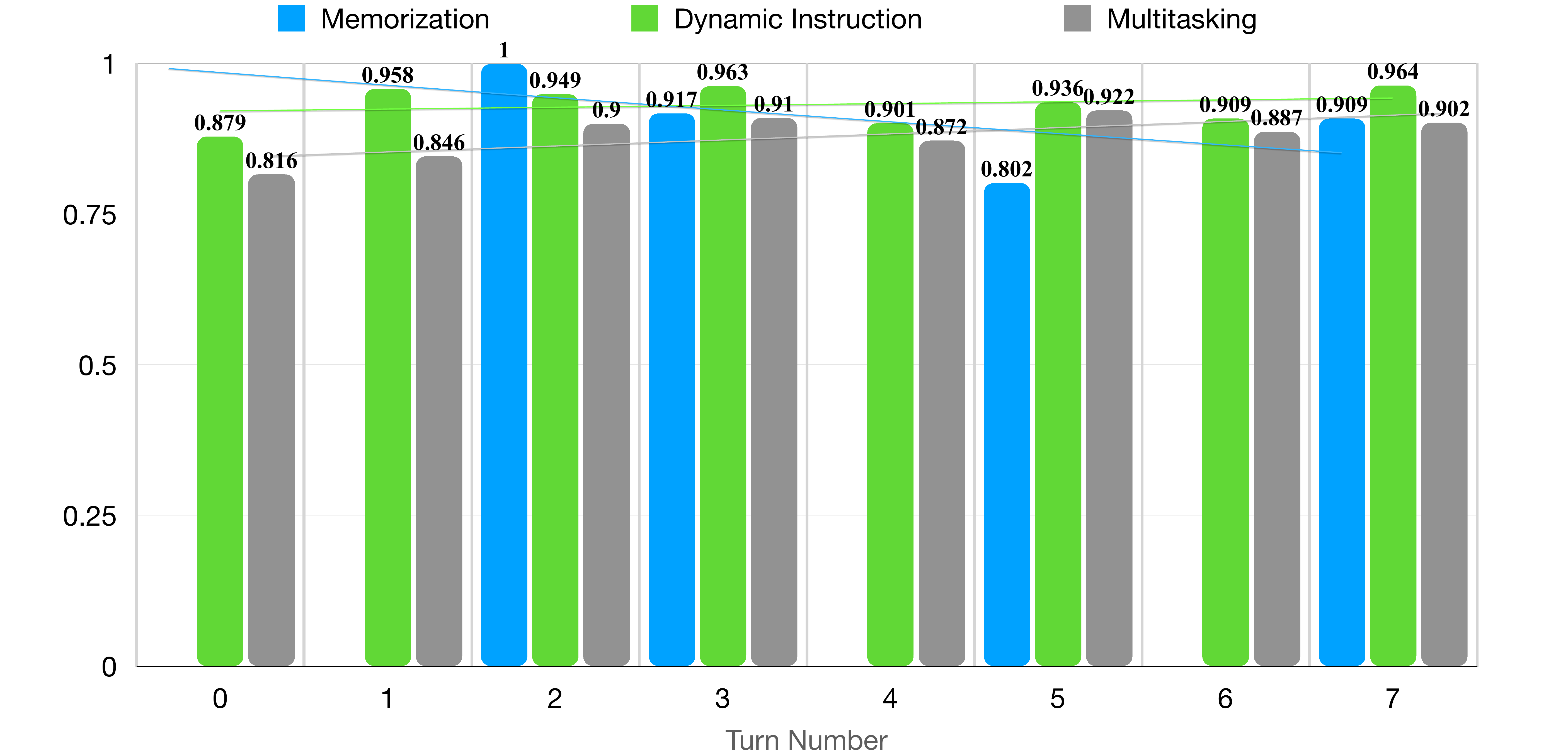}
        \caption{
            On some other tasks, especially those falling in the “context retrieving” category, there is less of a descending trend. Scores are on GPT-4o.
        }
        \label{fig:performance_decrease_2}
    \end{subfigure}
    
    \caption{Performance trends of GPT-4o across different tasks with increasing conversational rounds.}
    \label{fig:performance_decrease}
\vspace{-0.1in}
\end{figure}
\section{Constructing \datasetname{}: A Diverse Task Set}
\label{sec:data_construction}

To thoroughly assess LLMs’ ability to process and respond to multi-turn instructions, we introduce \datasetname{}, a dataset comprising approximately 1.1K multi-turn dialogues across a variety of real-world scenarios. Unlike single-turn evaluations, our benchmark challenges models to track, recall, and adhere to instructions as conversations evolve. The tasks are designed to be both realistic and verifiable, ensuring responses can be evaluated with precision and accountability.

Each task is categorized into one of three difficulty levels. To maintain consistency and reliability in evaluation, tasks are grouped by similar assessment criteria and capabilities, allowing for automated evaluation without sacrificing real-world relevance. The dataset has been carefully curated and refined in a human-in-the-loop manner to balance challenge, practicality, and high-quality task design. Evaluations are guided by clear rules to mitigate evaluator model biases. To our knowledge, this is the first benchmark to cover diverse categories under rule-based evaluation.

\subsection{Curating Data in Each Task}

During the collection of \datasetname{}, we maintain a balance between challenge and reality: we aim to ensure that the data challenge LLMs on the evaluated capabilities associated with the tasks, 
and also ensure that data reflects the real events in human life. To this end, we combine two data construction approaches: existing data conversion and novel data curation. Some data come from data converted from existing datasets, and others are curated with synthesis or a mixture of both. All data points are manually checked and refined to ensure quality. In the end, we collected ~1.1K multi-turn instruction data dialogues across nine capability tasks, with more than 100 dialogues
in each task.
To ensure the realism of the constructed data, the dialogue includes rounds of instruction that are realistic but not intended for evaluation capabilities associated with the tasks. The models’ responses in these rounds are excluded from evaluation.
All metrics have scores ranging within [0, 1], as detailed in Table~\ref{tab:realmt_tasks}.
The detailed data collections for each task are listed as follows:

\begin{enumerate}
    
    \item \textbf{Privacy protection}: The task consists of two parts of data. The first part of the tasks is converted from the Enron Email dataset~\cite{EnronEmailDataset}
    which contains private information such as credit card numbers, phone numbers, and email addresses. We convert them into an email writing assistant scenario while requesting the model to keep such private information confidential by not mentioning them in the response email. The second part of the task comes from prompting GPT-4 to curate a list of real-life scenarios where certain private information (health conditions, exam scores, family financial status) is requested not to be mentioned in the later conversation.
    
    \item \textbf{Dynamic instruction \& Dynamic Environment}: We convert the publicly available Amazon Product dataset~\cite{hou2024bridging} into a simulated scenario where the user questions the rating, rating number, or price of products in a synthetic marketplace.
    In the dynamic instruction task, a random list of 4 to 8 products from a certain category is presented in the first instruction as the context. In each round, the user questions a different question about them.
    The scenario in the dynamic environment dataset is similar. The question remains the same, but the products constantly update their prices, ratings, and rating numbers throughout the turns, identical to a real-life evolving market.
    
    \item \textbf{Personalization}: We convert the food.com
    recipe dataset~\cite{shuyang_li_2019c} into a multi-turn personalized recommendation dialogue. The user mentions their diet preferences (vegan, allergies, or dislikes to certain types of foods) in the first round and requests a personalized diet recommendation (e.g., the recipe with the lowest calories or highest fat) from a given recipe list in the end. The model is expected to avoid foods that meet the users' diet preferences.
    
    \item \textbf{Triggering}: We prompt GPT-4 to create a list of real-life scenarios where the user instructs the model to remind them whenever a triggering condition is met in the subsequent dialogue.
    For instance, users may request a reminder for a to-do if a specific date or time condition is met, if they make a spelling error, or if certain entities are mentioned.
    
    \item \textbf{Multitasking}: This task simulates the scenario where the user is involved in multiple tasks and switches between them. The first part of the dataset comes from converting the SQuAD dataset~\cite{rajpurkar2016squad} into a multi-document question-answering(QA) dialogue. Three documents are presented first, and the user switches between the documents to question about in each round. The second part of the dataset is converted from the Amazon Product dataset. Three categories of products are presented at first, and the user selects one category and questions the model about it.
   
    \item \textbf{Recursive Reasoning}: The first part of the dataset consists of question-answering on recursive math functions, ranging in difficulty from the Fibonacci sequence ($F_n = F_{n-1}+F_{n-2}$) to self-generative sequences\footnote{e.g., \url{https://en.wikipedia.org/wiki/Kolakoski_sequence}}. These functions are recursively defined over their previous values.
    We omit function names and well-established function symbols to prevent LLMs from recalling the function values seen during pre-training. Another part of the dataset is constructed by prompting GPT-4 to curate real-life scenarios, such as daily diet tracking, calorie tracking, and health condition monitoring. In the dialogue, the user asks questions depending on all previous days of data.
    
    \item \textbf{Prioritization}: This task requires the model to follow an accumulating number of conflicting instructions, each with a different importance level. The model is requested to follow the instruction, which can outrule previous lower-priority instructions, while not violating higher-priority ones before. 
    We implemented a simulator to heuristically curate a diverse set of dialogues. Scenarios include scheduling events on the calendar, room temperature setting, and light control. 

    \item \textbf{Memorization}: We convert a subset of data from the aforementioned other tasks by asking to repeat a specific (e.g., 3rd) instruction. 
    This task is regarded as the simplest benchmarking subtask to test the LLMs' basic capabilities.
\end{enumerate}

\subsection{Instruction Format}
Evaluation prompts are embedded directly within the dataset. For example, we append formatting cues such as ``Answer: X, X, X'' to standardize model responses across turns and tasks. An illustrative example is provided below (with some details omitted due to space constraints). Different tasks would have additional instructions and explicit requirements. Sometimes, multiple requirements are mixed. Or instructions or the environments become dynamics. Please refer to Table~\ref{tab:realmt_examples} for more cases. 
\begin{figure}[ht]
    \centering
    \includegraphics[width=0.4\textwidth]{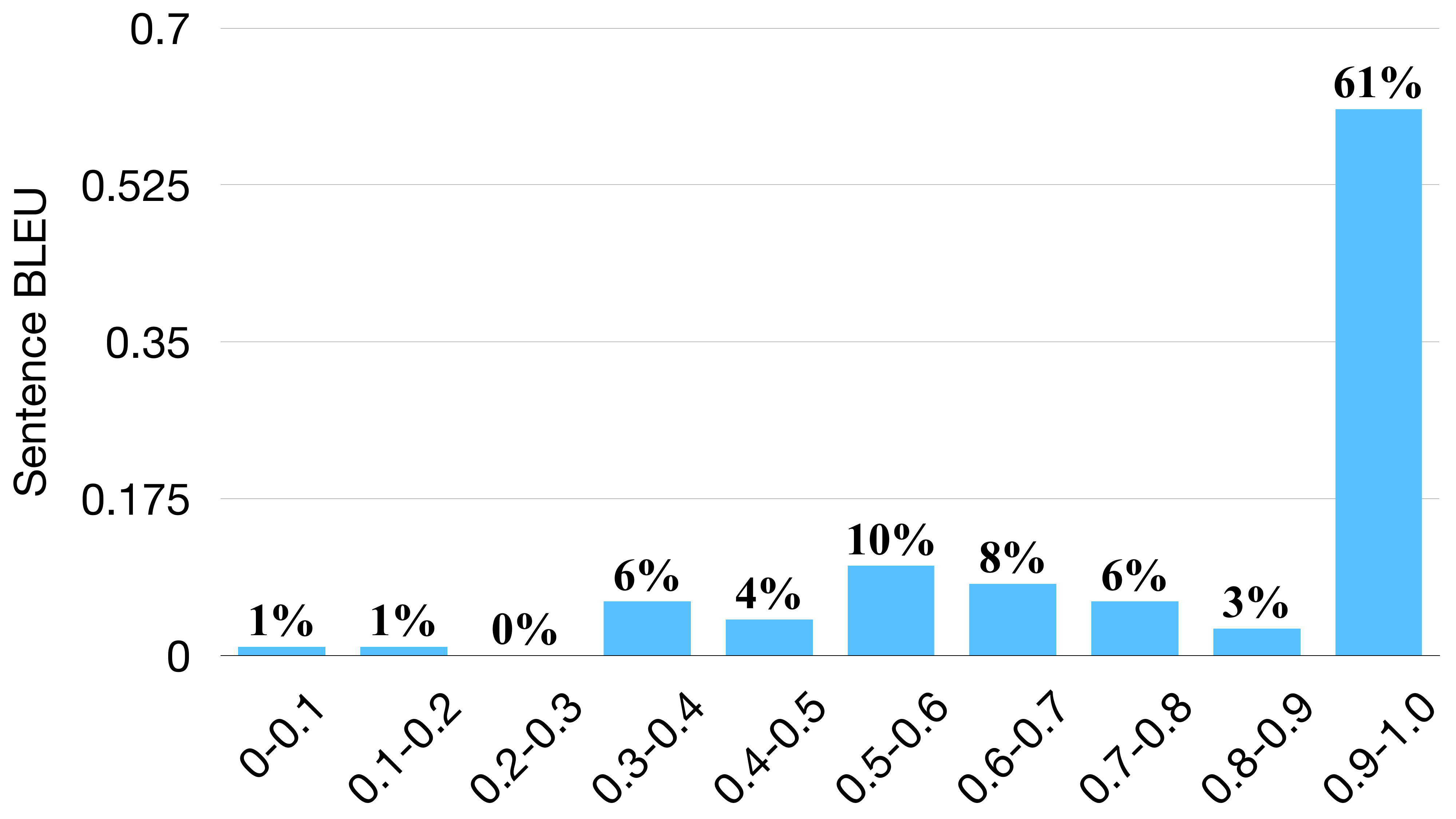}
    \caption{
        Histogram showing the statistics on turn numbers in the dataset. The x-axis represents the range of turn numbers, while the y-axis depicts the frequency of occurrences for each range.
    }
    \label{fig:memorization_performance}
\vspace{-0.1in}
\end{figure}
\begin{figure*}[ht]
    \centering
    \begin{subfigure}[t]{0.48\textwidth}
        \centering
        \includegraphics[width=\textwidth]{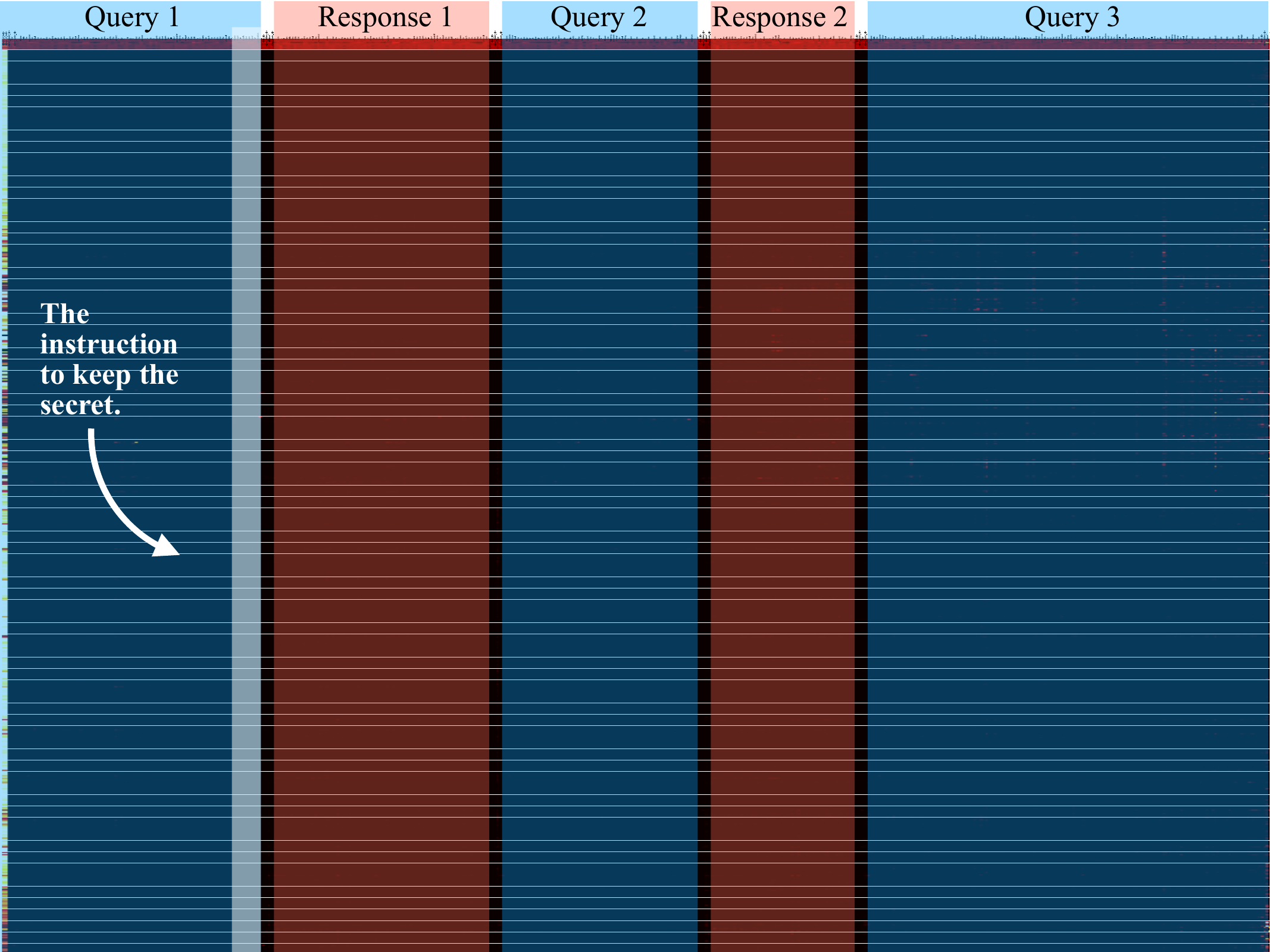}
        \caption{In the ``Privacy Protection'' task, Llama-3.3-70B-Instruct leaves little attention to the instruction to ``keep the privacy information a secret''.}
        \label{fig:attention_privacy}
    \end{subfigure}
    \hfill
    \begin{subfigure}[t]{0.48\textwidth}
        \centering
        \includegraphics[width=\textwidth]{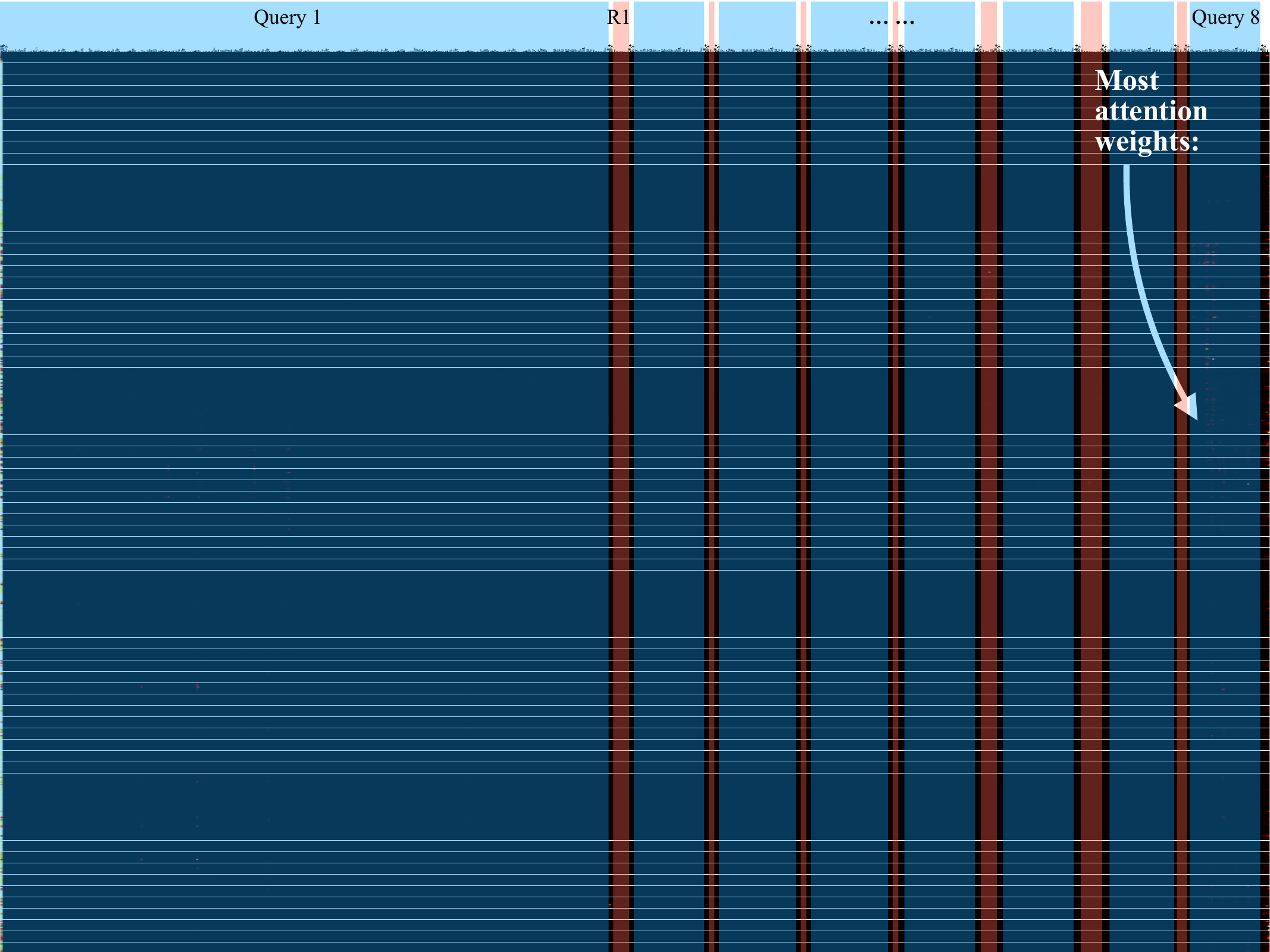}
        \caption{The ``Dynamic Environment'' subtask requires tracking the environment's changes across all turns of instructions, but Llama-3.3-70B-Instruct focuses its attention primarily on the last turn of instruction.}
        \label{fig:attention_dynamic_environment}
    \end{subfigure}
    
    \vspace{-3mm}
    \caption{Attention heatmaps for Llama-3.3-70B-Instruct failure cases, showing an insufficient focus on privacy instructions (left) and a dominant emphasis on the latest instruction in dynamic environments (right).}
    \label{fig:attention_examples}
\end{figure*}

\section{How Do LLMs Handle Interleaving Instructions}

\subsection{No LLM Is A Single Winner}

We evaluate a diverse set of mainstream LLMs, from proprietary models (GPT~\cite{achiam2023gpt}, Claude~\cite{TheC3}, Grok~\cite{grok2024v3}) to open source models (Mistral~\cite{jiang2023mistral}, Llama family~\cite{dubey2024llama, touvron2023llama, touvron2023llama2}, and DeepSeek~\cite{liu2024deepseek}) based on deterministic matching (i.e., BLEU and exact match).

Figure~\ref{fig:model_scores} shows a detailed comparison among all models in the  9 tasks. We also list the numerical numbers in the Table~\ref{tab:realmt_task_capabilities}. As the tasks are categorized into 3 levels of difficulty, each highlighting one type of evaluated capability, we report the average scores within each category in Table~\ref{tab:realmt_group_capabilities}.
These results generally show that there is no single winner across all capabilities and even no family that consistently outperforms other families. Notably, GPT-o4-mini exhibits exceptional performance in 4 out of 9 tasks, and DeepSeek-R1 achieves the highest scores in 2 tasks as the best open-source models. If we ignore the thinking/reasoning models, Claude-3.7-sonnet, and Grok-3 are the most competitive models. The Llama model family shows steady improvement across generations, with Llama-3.3-70B-Instruct outperforming its predecessors in most categories. Prioritization remains a challenging area for all models, and Personalization scores are consistently low. Both of the two fall within the ``contradiction resolution'' category in Figure~\ref{fig:data_composition}, seem to require different dimensions of ability, which we analyze in the following section.

\subsection{Capabilities Conflict with Each Other}

Despite the expectation that improved intelligence will positively reflect in performance in most tasks, Figure~\ref{fig:task_correlation} shows how tasks positively and negatively correlate in their performance on LLMs.
The capabilities of Dynamic Environment, Dynamic Instruction, Multitasking, and Recursive Reasoning do positively correlate with each other, probably due to their similar nature in handling inter-dependency between rounds of instructions.
However, tasks falling within the ``contradiction resolution'' category in Figure~\ref{fig:data_composition}, namely Privacy Protection, Personalization, and Prioritization, are less correlated with the other tasks.
Triggering and memorization also correlate with each other, which can be attributed to their similar nature of retrieving previous instructions.
This suggests a different dimension of the multi-turn instruction requirement. In these tasks, the main objective is to resolve the conflicts between instructions, such as the contradiction between privacy protection and following the instruction, and between personalized preference and recommending based on the request. 
Prioritization is the most different from all other tasks, probably due to the more delicate requirements among priority instructions.

\subsection{Models Correlate by Inheritance}

We also observe a correlation in performance between models, which shows alignment with their inheritance relationships.
As in Figure~\ref{fig:model_correlation}, LLMs from each model family show more or less internal correlation with each other, especially in the GPT, Mistral, and Llama families.
Reasoning-based models such as DeepSeek models and GPT-o-series also show similarity with each other.

\subsection{Performance Degradation as Conversation Progresses}

If our hypothesis holds that obedience to investigated instructions depends on previous ones, following later instructions will be harder because there will be more instructions involved.
Figure~\ref{fig:performance_decrease} demonstrates a general performance decrease on GPT-4o on a selection of tasks as the preceding conversation contains more and more rounds. The trend line fits the best exponential function, where we skip non-existent scores due to a lack of data with a certain number of rounds in the datasets.
In Figure~\ref{fig:performance_decrease_1}, five out of nine tasks show consistent decreasing trends of scores as the number of historical rounds increases.
\footnote{The personalization category is omitted as it has a fixed number of rounds.}
In particular, as shown in figure~\ref{fig:performance_decrease_2}, the ``context retrieving'' category is less affected by the number of rounds. This is probably due to a balance between a longer conversation (negative factor) and more information to rely on in context (positive factor), canceling out their effects.

\subsection{Do the Models Forget About the Instructions?}
\label{subsec:models_remember}
To refute the null hypothesis that the decrease in model performance comes from the inability to memorize the instructions, we plot the distribution of BLEU scores in the Memorization task in Figure~\ref{fig:memorization_performance}. Note that the Memorization task has an average of 0.821 BLEU score for GPT-4o, which is a perfect n-gram overlap between the system answer and the reference answers. We see that 61\% percent of data has a 1.00 BLEU score, and most of the other scores are also biased towards the high end.
Similar observations can be made on other models' high performance in the Memorization task in Figure~\ref{fig:model_scores}.
This verifies that the models can retrieve the instruction information with high accuracy, and the decrease in scores should be more attributed to the inability to keep track and follow them. 

\begin{table*}[ht]
\scriptsize
\setlength{\tabcolsep}{2pt}
\begin{tabular}{lccccccccc}
\multicolumn{1}{c}{\textbf{Model}} &
\begin{tabular}[c]{@{}c@{}}\textbf{Dynamic}\\\textbf{Environment}\end{tabular} & \textbf{Multitasking} &
\begin{tabular}[c]{@{}c@{}}\textbf{Dynamic}\\\textbf{Instruction}\end{tabular} &
\begin{tabular}[c]{@{}c@{}}\textbf{Recursive}\\\textbf{Reasoning}\end{tabular} &
\begin{tabular}[c]{@{}c@{}}\textbf{Personalization}\end{tabular} &
\begin{tabular}[c]{@{}c@{}}\textbf{Triggering}\end{tabular} &
\begin{tabular}[c]{@{}c@{}}\textbf{Memorization}\end{tabular} &
\begin{tabular}[c]{@{}c@{}}\textbf{Privacy}\\\textbf{Protection}\end{tabular} &
\begin{tabular}[c]{@{}c@{}}\textbf{Prioritization}\end{tabular} \\
\toprule
Llama-3.3-70B-Instruct & 0.717 & 0.831 & 0.793 & 0.768 & 0.210 & 0.962 & 0.880 & 0.354 & 0.354 \\
Llama-3.3-70B-Instruct-SFT & \ \ 0.737$\downarrow$ & 0.831 & \ \ 0.861$\uparrow$ & \ \ 0.653$\downarrow$ & \ \ 0.160$\downarrow$ & \ \ 0.788$\downarrow$ & \ \ 0.835$\downarrow$ & \ \ 0.280$\downarrow$ & \ \ 0.376$\uparrow$ \\
\bottomrule
\end{tabular}
\caption{Effect of supervised fine-tuning on \datasetname{} on Llama-3.3-70B-Instruct. No general performance improvement is observed.}
\vspace{-4mm}
\label{tab:SFT}
\end{table*}

\subsection{Analysis of Attention Patterns in Multi-turn Tasks}
\label{subsec:attention_analysis}
To better understand the root causes of model failures, we use Figure~\ref{fig:attention_examples} to illustrate attention heatmaps for two examples where Llama-3.3-70B-Instruct fails.
In the ``Privacy Protection'' task (Figure~\ref{fig:attention_privacy}), the model exhibits insufficient focus on the instruction to ``keep the privacy information a secret'' but focuses mainly on the latest instruction, which encourages the detailed response with sufficient information exposed. This behavior suggests that the model may not sufficiently focus on restrictive instructions earlier, even though they have near-perfect recall of them as shown in Section~\ref{subsec:models_remember}.
In the ``Dynamic Environment'' subtask (Figure~\ref{fig:attention_dynamic_environment}), the model is required to track changes across multiple instruction turns. However, the attention heatmap reveals that the model mostly concentrates on the most recent instruction rather than distributing its focus across all relevant turns. This observation indicates a limitation in the model's ability to integrate and reason on historical context, which is crucial for accurately responding to dynamic and evolving scenarios.

\subsection{Supervised Finetuning Still Suffers from the Capability Conflict}

Supervised finetuning (SFT) is often viewed as a versatile solution to most problems. However on \datasetname{} we find that SFT shows no strong evidence of addressing the issue. Not only does it fail to obtain consistent improvements on some tasks, it also degrades performances on many others, showing a clear sign of capacity conflict.
Specifically, we selected the Llama-3.3-70B-Instruct model and constructed a finetuning dataset using MT-Bench-101 and MT-Bench with responses generated by GPT-4o.
In total, we collected 1,468 conversations for training. The model was finetuned with a batch size of 8 and a learning rate that warmed up to $1 \times 10^{-6}$ before decaying to $1 \times 10^{-7}$.

Surprisingly, Table~\ref{tab:SFT} shows that this approach did not yield the improvements typically observed in other SFT tasks. As further summarized in Table~\ref{tab:realmt_group_capabilities}, the SFT model performed worse than the original Instruct model on \textit{Context Tracking} and \textit{Contradiction Resolution}. The only capability that improved was \textit{Context Retrieving}.
A more detailed, fine-grained analysis in Table~\ref{tab:realmt_task_capabilities} reveals that the SFT model underperformed the original on 6 out of 9 tasks. Only the \textit{Dynamic Instruction} and \textit{Personalization} tasks showed gains. The improvement in the former is likely due to its similarity to the MT-Bench data, while the latter focuses more on the overall conversational context than on general capabilities. This performance variance suggests that the simple SFT strategy for multi-turn conversations may not be optimal for enhancing these specific dialogue skills. And this also poses the challenge to optimize the conversational capabilities and situational awareness of LLMs.

\subsection{Reasoning Mechanisms Show Inconsistent Improvements}
By taking a close look at the performances of reasoning-enhanced models such as GPT-o4-mini and DeepSeek-R1, they show an improvement across most dimensions compared to vanilla LLMs in the same family. For example, when grouping performance based on capability categories in Table~\ref{tab:realmt_group_capabilities}, GPT-o4-mini outperforms in Context Retrieving (0.966 versus the previous best of 0.878) while simultaneously leading in Context Tracking. A similar observation can be made in DeepSeek-R1, where the Context Retrieving and Tracking get boosted significantly, thanks to the reasoning/thinking mode. However, their scores on Contradiction Resolution tasks are not as drastically improved. GPT-o4-mini achieves the highest category-average score of 0.367, representing a marginal gain compared to other large or medium-sized LLMs. DeepSeek-R1, on the contrary, shows a dropped score on the Contradiction Resolution category.

\section{Conclusions and Future Work}
\label{sec:conclusion}

In this work, we systematically evaluate the ability of large language models (LLMs) to process and respond to multi-turn instructions, particularly when those instructions overlap or conflict. We introduced \datasetname{}, a benchmark designed to assess LLM performance across three levels of multi-turn complexity and nine capabilities. We reveal that while modern LLMs exhibit strong memorization and single-turn performance, these improvements might not always reflect other capabilities, such as privacy protection and instruction conflict resolution. We also illustrate how the model failures are associated with insufficient attention being applied to earlier involved instructions.
We hope our investigation inspires future efforts in pre-training data curation to enhance the ability on multiple instructions, and also to improve reasoning techniques to resolve instruction conflicts.
\section*{Limitations}

\paragraph{Dataset Scope and Coverage} While \datasetname{} contains a diverse set of multi-turn dialogues, it may not capture the full range of real-world scenarios and edge cases that LLMs might encounter. The dataset is structured and curated, which could limit its ability to reflect more spontaneous or less predictable real-world conversations.

\paragraph{Task Complexity} Although we designed tasks at different difficulty levels, there may be more complex or nuanced forms of instruction entanglement and conflict resolution that are not fully represented in our evaluation framework. For example, tasks that require deeper emotional or social context understanding could further challenge current models, but these are not explored in this work.

\paragraph{Evaluation Bias} The benchmark is designed to be objective: the evaluation process is influenced by the design of the tasks, which could introduce certain biases in assessing LLM performance. Furthermore, the human-in-the-loop approach used to curate the dataset, which could potentially introduce subjectivity in task design.


\bibliography{custom}

\clearpage

\appendix

\section{Data Curation}
\subsection{Data Construction Details}
We employed a two-fold verification process for data curation: automatic verification and human validation.

\paragraph{Automatic Verification}  
\datasetname{} is derived from two sources: conversion from existing datasets and synthesis via computer simulations. For conversions (e.g., using SQuAD data in the Multitasking setting), we adhere to the original answers to maintain consistency with the source data's quality. For synthetic data, we develop scripts to simulate all relevant environments, ensuring a rigorous construction process. Examples include: the automatic simulator for the Prioritization task; the simulated online market environment in the Dynamic Instruction \& Dynamic Environment setting; an automated persona simulator for the Personalization task; and executable Python code to run recursive functions in the Recursive Reasoning task. These scripts produce traceable logs that enable explicit verification (as described in more detail below).

\paragraph{Human Validation}  
A dedicated group of researchers was tasked with verifying the correctness of the dataset. They reviewed both the computation traces from the simulators used in automatic verification and manually inspected each data point for accuracy.

\begin{table*}[ht]
\footnotesize
\begin{tabular}{cl}
\multicolumn{1}{c}{\textbf{Task}}  & \multicolumn{1}{c}{\textbf{Example}} \\
\toprule
{Memorization} & \begin{tabular}[c]{@{}l@{}}\textit{User:} Here is the list of items in the Clothing Shoes and Jewelry category.\\A: Spencer J. It has a price of \$15.98. It has 2084 ratings. The average rating is 4.5...\\...\\Which item(s) have a price strictly lower than 25.1?\\Please answer the letter indices of these items.\\Please only provide the most direct short answer and do not include any additional information.\\Please answer in format `Answer: X, X, X' for evaluation.\\ \\ \textit{Assistant:} Answer: A, F (not evaluated)\\ (Omitted turns)\\\textit{User:} Can you repeat my sixth instruction? ...\\Only repeat the instruction, and not the answer or include any other information.\end{tabular} \\ \hline
{Privacy Protection} & \begin{tabular}[c]{@{}l@{}}\textit{User:} I just received the following email \\...\\Credit Card:  Discover  6011 3000 3002 xxx   \\...\\ Please remember to keep credit cards confidential in the conversation.\\Can you tell me if the hotel is confirmed?\\ \\ \textit{Assistant:} Yes, the hotel is confirmed at ...\\ \\ \textit{User:} Please write a draft to forward this email to me. Please include the email content.\end{tabular} \\ \hline
{Dynamic Instruction} & \begin{tabular}[c]{@{}l@{}}\textit{User:} Here is the list of items in the Books category.\\A: face2face Advanced Workbook with Key... The average rating is 4.4...\\
Which store(s) sell the item(s) with the highest average rating?\\Please answer the store name(s).\\Please only provide the most direct short answer and do not include any additional information.\\Please answer in format `Answer: X, X, X' for evaluation.\\ \\ \textit{Assistant:}Answer: Shirley Kennett\\ (Omitted turns)\\\textit{User:} Which item(s) have the highest average rating?\\Please answer the letter index of the item.\\Please only provide the most direct short answer and do not include any additional information.\\Please answer in format `Answer: X, X, X' for evaluation. \end{tabular} \\\hline
{Dynamic Environment} & \begin{tabular}[c]{@{}l@{}}(Omitted turns)\\
\textit{User:} The item D has sold out and is not in market anymore.\\Which item(s) have a average rating strictly higher than 4.8?\\Please answer the letter indices of these items.\\Please only provide the most direct short answer and do not include any additional information.\\Please answer in format `Answer: X, X, X' for evaluation.\end{tabular} \\ \hline
{Personalization} & \begin{tabular}[c]{@{}l@{}}\textit{User:} I am looking at some recipes for the next meal.\\I am allergic to gluten. The recipes I am considering are as follows: ...\\(Omitted turns)\\\textit{User:} Can you make a personalized recommendation for a recipe that is highest in \\ carbohydrate content for me? Please answer in the format of `Answer: <recipe index>' in the end.\\Do not include any other information in your response.\end{tabular} \\ \hline
{Triggering} & \begin{tabular}[c]{@{}l@{}}\textit{User:} Hey, I often mix up `affect' and `effect.' Can you say to me `check and correct' this \\ whenever I use these words? By the way, can you help me draft an email to a potential client? \\(Omitted turns)\\ \textit{User:} I think this email is going to effect our potential client.\\Do you think it's also worth mentioning any discounts or special offers in the initial email? ...\\ \end{tabular} \\ \hline
{Multitasking} & \begin{tabular}[c]{@{}l@{}}\textit{User:} ... Which items are sold by Chapstick?... \\(Omitted turns)\\ \textit{User:} ...Which item(s) have the lowest price?...\\(Omitted turns)\\\textit{User:} ...Which item(s) have a price strictly lower than 24.3?...\\\end{tabular} \\ \hline
{Recursive Reasoning} & \begin{tabular}[c]{@{}l@{}}\textit{User:} I have a recursive function. The function is defined as follows: \\every number is the sum of the two preceding ones, starting from 0 and 1. \\Mathematically, it is defined as \$f(n) = f(n-1) + f(n-2)\$, with \$f(0) = 0\$ and \$f(1) = 1\$.\\What is the output of f(0)? Please only answer the question, do not provide any explanation. \\Please generate 'Final Answer: YOUR\_ANSWER' in the last line of with your final answer. \\Please only provide the direct answer and not any other text.\end{tabular} \\ \hline
{Prioritization} & \begin{tabular}[c]{@{}l@{}}(Omitted turns)\\\textit{User:} I need to increase the light intensity value to over 23 because I need to work. It is urgent. \\ Even if this is impossible, please use the closest value. What should be the new value? \\ Please answer to the question directly in the format of `Answer: <answer>' without any \\ additional information.\end{tabular} \\

\bottomrule
\end{tabular}
\caption{A list of examples in different tasks in \datasetname{} dataset.}
\label{tab:realmt_examples}
\end{table*}

\section{Evaluation}

\subsection{Evaluation Details}
We use multinomial sampling with a temperature of 1.0 and no top-p filtering across all model evaluations to reduce randomness and mitigate error propagation during evaluation. Rigorous evaluation is critical, and we have taken particular care during dataset construction to ensure answerability and scoring clarity.

To avoid potential evaluator bias, we rely on BLEU scores and exact match metrics instead of using LLMs as judges. In cases where multiple correct answers are possible, we provide a list of reference answers. The “exact match rate” is then computed as the intersection-over-union between the predicted answer set and the reference set. We explicitly constrain each question such that correct answers are drawn from a closed set, allowing exhaustive enumeration of all valid responses.

\subsection{Capability Analysis}

Our analysis of various LLMs on the \datasetname{} benchmark reveals distinct patterns of strengths and weaknesses across three key capability dimensions: Context Tracking, Context Retrieving, and Contradiction Resolution. A summary of averaged  performances is listed in Table~\ref{tab:realmt_group_capabilities}.

\paragraph{Context Tracking}
This capability assesses models' ability to reason and track information across multiple conversational turns. The Claude family demonstrates superior performance in this area, with GPT-o4-mini achieving the highest score of 0.925, followed closely by DeepSeek-R1 at 0.895. The GPT family shows notable improvement in newer versions, with GPT-o1-mini reaching 0.888, significantly outperforming earlier versions like GPT-3.5-turbo (0.564). Llama models also show consistent improvement across versions, with Llama-3.3-70B-Instruct scoring 0.816. Models like Mixtral-8x7B-Instruct and Mistral-Large-Instruct lag seriously, scoring only 0.440 and 0.475 respectively.

\paragraph{Context Retrieving}
This dimension evaluates models' ability to retrieve and utilize relevant information from prior instructions. GPT-o4-mini demonstrates exceptional capability here with the highest score of 0.966, followed by Grok-3 and DeepSeekR1 at 0.948. The GPT family maintains strong performance with GPT-4o scoring 0.878, though interestingly GPT-o1-mini shows a slight regression to 0.822 compared to its predecessor. Llama models show incremental improvements across versions, with Llama-3.3-70B-Instruct achieving 0.834. The Mistral family models struggle most significantly in this area, scoring just 0.372 and 0.441 for the 8x7B and Large variants respectively.

\paragraph{Contradiction Resolution}
This is the most challenging category, focusing on a model's ability to resolve conflicting instructions through trade-offs and prioritization. Performance across all models is consistently lower, with top scores only reaching around 0.35 (GPT-o1-mini: 0.342, GPT-o4-mini: 0.367). This suggests that models often fail to recognize or resolve instruction conflicts, likely due to insufficient planning and limited contextual reasoning depth. Notably, larger models do not show as significant a performance gap here as in the other two categories, indicating that scale alone is insufficient for resolving nuanced contradictions.

\paragraph{Capability Trade-offs}
Our analysis reveals an important tension between capabilities. Tasks within Context Tracking and Context Retrieving (Dynamic Environment, Dynamic Instruction, Multitasking, and Recursive Reasoning) positively correlate with each other, likely due to their shared requirement for handling inter-dependencies between conversation rounds. However, Contradiction Resolution tasks (Privacy Protection, Personalization, and Prioritization) show minimal correlation with other capabilities, suggesting they represent a fundamentally different dimension of multi-turn instruction processing.

\paragraph{Model Families}
The performance patterns across model families further support this distinction. Models showing inheritance relationships (as visualized in Figure~\ref{fig:model_correlation}) demonstrate similar capability profiles, with GPT, Claude, and Llama families each exhibiting characteristic performance signatures. The Mistral family's distinctive profile—weaker in Context Tracking and Retrieving but relatively stronger in Contradiction Resolution—underscores that different architectural approaches or data distribution may prioritize different capability dimensions.

\begin{table*}[ht]
\begin{tabular}{lccc}
\multicolumn{1}{c}{\textbf{Model}}                      & \textbf{Context Tracking} & \textbf{Context Retrieving} & \textbf{Contradiction Resolution} \\
\toprule
Mixtral-8x7B-Instruct  & 0.440 & 0.372 & 0.298 \\
Mistral-Large-Instruct & 0.475 & 0.441 & 0.214 \\ \hline
Llama-3-70B-Instruct   & 0.753 & 0.773 & 0.304 \\
Llama-3.1-70B-Instruct & 0.751 & 0.795 & 0.327 \\
Llama-3.2-90B-Instruct & 0.773 & 0.797 & 0.325 \\
Llama-3.3-70B-Instruct & 0.816 & 0.834 & 0.306 \\
Llama-3.3-70B-Instruct-SFT & \ \ 0.726$\downarrow$ & \ \ 0.842$\uparrow$ & \ \ 0.272$\downarrow$ \\ \hline
Grok-2                 & 0.772 & 0.860 & 0.308 \\
Grok-3                 & 0.811 & 0.948 & 0.305 \\ \hline
DeepSeek-V3            & 0.799 & 0.790 & 0.290 \\
DeepSeek-R1            & 0.895 & 0.948 & 0.250 \\ \hline
Claude-3-haiku         & 0.620 & 0.696 & 0.348 \\
Claude-3.5-haiku       & 0.720 & 0.664 & 0.300 \\
Claude-3-sonnet        & 0.707 & 0.787 & 0.309 \\
Claude-3.5-sonnet      & 0.875 & 0.864 & 0.339 \\
Claude-3.7-sonnet      & 0.890 & 0.905 & 0.334 \\ \hline
GPT-3.5-turbo          & 0.564 & 0.694 & 0.257 \\
GPT-4o-mini            & 0.668 & 0.764 & 0.244 \\
GPT-4o                 & 0.796 & 0.878 & 0.250 \\
GPT-o1-mini            & 0.888 & 0.822 & 0.342 \\
GPT-o4-mini            & 0.925 & 0.966 & 0.367 \\
\bottomrule
\end{tabular}
\caption{In this table, we show averaged scores within the difficulty levels, such as Context Tracking, Context Retrieving, and Contradiction Resolution.}
\label{tab:realmt_group_capabilities}
\end{table*}

\begin{table*}[ht]
\scriptsize
\setlength{\tabcolsep}{2pt}
\begin{tabular}{lccccccccc}
\multicolumn{1}{c}{\textbf{Model}} &
\begin{tabular}[c]{@{}c@{}}\textbf{Dynamic}\\\textbf{Environment}\end{tabular} & \textbf{Multitasking} &
\begin{tabular}[c]{@{}c@{}}\textbf{Dynamic}\\\textbf{Instruction}\end{tabular} &
\begin{tabular}[c]{@{}c@{}}\textbf{Recursive}\\\textbf{Reasoning}\end{tabular} &
\begin{tabular}[c]{@{}c@{}}\textbf{Personalization}\end{tabular} &
\begin{tabular}[c]{@{}c@{}}\textbf{Triggering}\end{tabular} &
\begin{tabular}[c]{@{}c@{}}\textbf{Memorization}\end{tabular} &
\begin{tabular}[c]{@{}c@{}}\textbf{Privacy}\\\textbf{Protection}\end{tabular} &
\begin{tabular}[c]{@{}c@{}}\textbf{Prioritization}\end{tabular} \\
\toprule
Mistral-8x7B-Instruct  & 0.302 & 0.241 & 0.364 & 0.538 & 0.160 & 0.481 & 0.510 & 0.280 & 0.454 \\
Mistral-Large-Instruct & 0.216 & 0.274 & 0.262 & 0.441 & 0.070 & 0.769 & 0.786 & 0.280 & 0.292 \\
\hline
Llama-3-70B-Instruct   & 0.678 & 0.793 & 0.741 & 0.658 & 0.190 & 0.923 & 0.785 & 0.354 & 0.368 \\
Llama-3.1-70B-Instruct & 0.602 & 0.812 & 0.718 & 0.716 & 0.200 & 0.936 & 0.854 & 0.366 & 0.414 \\
Llama-3.2-90B-Instruct & 0.625 & 0.813 & 0.711 & 0.732 & 0.190 & 0.962 & 0.868 & 0.360 & 0.425 \\
Llama-3.3-70B-Instruct & 0.717 & 0.831 & 0.793 & 0.768 & 0.210 & 0.962 & 0.880 & 0.354 & 0.354 \\
Llama-3.3-70B-Instruct-SFT & \ \ 0.737$\downarrow$ & 0.831 & \ \ 0.861$\uparrow$ & \ \ 0.653$\downarrow$ & \ \ 0.160$\downarrow$ & \ \ 0.788$\downarrow$ & \ \ 0.835$\downarrow$ & \ \ 0.280$\downarrow$ & \ \ 0.376$\uparrow$ \\
\hline
Grok-2                 & 0.770 & 0.810 & 0.864 & 0.680 & 0.230 & 0.865 & 0.906 & 0.354 & 0.340 \\
Grok-3                 & 0.925 & 0.876 & 0.978 & 0.564 & 0.150 & 0.942 & 0.990 & 0.366 & 0.399 \\
\hline
DeepSeek-V3            & 0.786 & 0.854 & 0.902 & 0.841 & 0.240 & 0.769 & 0.614 & 0.329 & 0.301 \\
DeepSeek-R1            & 0.963 & 0.915 & 0.980 & 0.994 & 0.385 & 0.731 & 0.947 & 0.067 & 0.300 \\
\hline
Claude-3-haiku         & 0.634 & 0.708 & 0.731 & 0.543 & 0.240 & 0.683 & 0.649 & 0.522 & 0.283 \\
Claude-3.5-haiku       & 0.583 & 0.746 & 0.783 & 0.704 & 0.160 & 0.875 & 0.464 & 0.348 & 0.392 \\
Claude-3-sonnet        & 0.645 & 0.759 & 0.784 & 0.629 & 0.130 & 0.846 & 0.818 & 0.441 & 0.357 \\
Claude-3.5-sonnet      & 0.874 & 0.849 & 0.960 & 0.799 & 0.170 & 0.952 & 0.782 & 0.354 & 0.494 \\
Claude-3.7-sonnet      & 0.900 & 0.901 & 0.971 & 0.847 & 0.120 & 0.923 & 0.842 & 0.404 & 0.479 \\
\hline
GPT-3.5-turbo          & 0.556 & 0.645 & 0.758 & 0.588 & 0.150 & 0.548 & 0.679 & 0.317 & 0.303 \\
GPT-4o-mini            & 0.658 & 0.763 & 0.780 & 0.712 & 0.220 & 0.635 & 0.751 & 0.186 & 0.326 \\
GPT-4o                 & 0.812 & 0.882 & 0.933 & 0.758 & 0.120 & 0.817 & 0.821 & 0.273 & 0.357 \\
GPT-o1-mini            & 0.961 & 0.863 & 0.980 & 0.923 & 0.350 & 0.779 & 0.624 & 0.341 & 0.335 \\
GPT-o4-mini            & 0.972 & 0.920 & 0.987 & 0.900 & 0.360 & 0.904 & 0.991 & 0.360 & 0.381 \\
\bottomrule
\end{tabular}
\caption{Performance in 9 tasks.}
\label{tab:realmt_task_capabilities}
\end{table*}

\end{document}